\documentclass[10pt,conference]{IEEEtran}

\usepackage{amsfonts}
\usepackage{graphicx,times,amsmath}
\usepackage{latexsym}
\usepackage{amssymb}
\usepackage{color}
\usepackage{url}  

\usepackage{booktabs}
\usepackage{multirow}
\usepackage{fancyhdr}

\usepackage{tikz}
\usetikzlibrary{shapes.geometric, arrows.meta, positioning, calc, backgrounds}

\tikzset{
    vertex/.style={
        circle, 
        draw=black, 
        fill=white, 
        thick, 
        minimum size=6mm, 
        inner sep=0pt,
        font=\bfseries\small
    },
    edge/.style={
        draw, 
        thick, 
        black
    },
    new_edge/.style={
        draw, 
        ultra thick, 
        green!60!black
    },
    del_edge/.style={
        draw, 
        ultra thick, 
        red, 
        dashed
    },
    op_label/.style={
        font=\bfseries\tiny,
        align=center,
        anchor=north,
        text width=3.5cm,    
        inner sep=1pt
    }
}

\IEEEoverridecommandlockouts

\definecolor {grey}{rgb}{0.2,0.2,0.2}
\definecolor {dkgr}{rgb}{0,0.3,0}

\title{Evolutionary Refinement of Generative Graph Topologies: A Hybrid WGAN-GA Approach}

\author{\IEEEauthorblockN{James Sargant}
\IEEEauthorblockA{Computer Science \\
Brock University, Canada\\
js17sy@brocku.ca}
\and
\IEEEauthorblockN{Seyedeh Ava Razi Razavi}
\IEEEauthorblockA{Computer Science \\
Brock University, Canada\\
ua23vc@brocku.ca}
\and
\IEEEauthorblockN{Renata Dividino}
\IEEEauthorblockA{Computer Science \\
Brock University, Canada\\
rdividino@brocku.ca}
\and
\IEEEauthorblockN{Sheridan Houghten}
\IEEEauthorblockA{Computer Science \\
Brock University, Canada\\
shoughten@brocku.ca}
}

\begin{document}

\maketitle

\begin{abstract}
Generating realistic graph-structured data is challenging due to discrete connectivity, varying graph sizes, and class-specific structural patterns. Recent Generative Adversarial Networks (GAN)-based graph generation methods improve edge modelling by learning connectivity and matching class-specific density distributions. However these models still exhibit noticeable deviations such as in degree and spectral distribution when compared to real graphs, indicating that important structural properties are not fully preserved. This work aims to reduce these deviations by refining the graphs produced by an existing GAN-based graph generator framework with a Genetic Algorithm (GA). In the GAN framework, the generator produces both node features and connectivity patterns, while a GNN-based critic evaluates graph realism and class consistency to ensure global structural and class alignment. Building on this foundation, we apply a GA to refine the edges of generated graphs. The refinement process guides synthetic graphs toward closer agreement with real data, while preserving diversity and novelty. Experimental results show that the GA refinement consistently lowers combined Maximum Mean Discrepancy (MMD) compared to the base model, leading to graphs that more closely match real structural patterns. This demonstrates that evolutionary refinement is an effective and flexible way to correct residual structural deviations in GAN-based graph generators, improving their suitability for realistic graph synthesis and data augmentation.

\end{abstract}

\section{Introduction}
\label{sec:intro}

Graph-structured data arises in many application areas, including social networks, molecular graphs, biological systems, and communication networks, where nodes and edges capture important information about underlying processes and system properties. The generation of realistic synthetic graphs has become increasingly relevant for applications such as privacy-preserving data sharing and data augmentation for graph-based machine learning models. However, generating graphs introduces fundamental challenges that differ from those of traditional data synthesis in Euclidean spaces. Unlike images or text, which have regular grid structures and fixed dimensionality, graphs exhibit irregular topologies, variable sizes, and no canonical node ordering. As a result, effective graph generation methods must learn complex dependencies between graph structure and node attributes while preserving the structural properties that define different graph types.

Early deep learning approaches to graph generation have explored various strategies to address these challenges. Methods such as DeepGMG~\cite{li2018learning} and GraphRNN~\cite{you2018graph} model graphs through sequential generation processes, while GraphVAE~\cite{simonovsky2018graphvae} leverages variational autoencoders to learn latent graph representations. Generative Adversarial Networks (GANs) have shown particular promise, with GraphGAN~\cite{wang2018graphgan}, EGraphGAN~\cite{wang2024graph}, and MolGAN~\cite{de2018molgan} demonstrating the potential of adversarial training for graph synthesis. LGGAN~\cite{fan2019labeled} further advanced the field by introducing labeled graph generation with comprehensive evaluation metrics including Maximum Mean Discrepancy (MMD) over degree, clustering, and orbit statistics. However, these GAN-based approaches often suffer from training instability, mode collapse, and difficulties in generating graphs with variable sizes.

The introduction of Wasserstein GANs (WGANs)~\cite{arjovsky2017wasserstein} marked a significant advancement in addressing the instability issues inherent in standard GANs. WGAN-GP~\cite{gulrajani2017improved} further improved upon this by enforcing the Lipschitz constraint through gradient penalty rather than weight clipping, enabling more effective training and better sample diversity. Recent work has successfully adapted the WGAN framework to graph generation by combining it with Graph Neural Networks (GNNs) as critics. In~\cite{ava_cascon, ava_arxiv}, a generator produces both node features and connectivity patterns, while a GNN-based critic evaluates graph realism and class consistency. These methods demonstrated superior performance in generating class-conditional graphs with improved training stability. However, despite these advances, these works show that these models still exhibit noticeable deviations such as in the degree spectral distribution when compared to real graphs. This limitation results in graphs with unrealistic connectivity patterns that fail to capture the complex structural dependencies between nodes and do not reflect the learned node feature distributions.

We address these core limitations by extending existing graph generation approaches with a Genetic Algorithm (GA)–based refinement stage. After the graph generator is fully trained, we apply a GA to refine the generated graphs through crossover and mutation operations that directly alter graph structure, such as edge presence and local connectivity patterns. The refinement process is guided by fitness measures derived from structural statistics of real graphs, with a particular focus on correcting deviations in degree, clustering coefficient and spectral distributions. 

Central to our refinement strategy is the concept of evolutionary edge editing, where graph topologies are iteratively optimized through a sequence of discrete structural modifications. This approach builds upon established generative representations that utilize elementary operations, such as adding, removing, or toggling edges, to evolve networks toward specific structural or functional targets~\cite{7529329, CEC19}. By leveraging these direct edge manipulations, the evolutionary stage can enforce precise local constraints and correct statistical deviations that are often smoothed over by the continuous approximations inherent in deep generative models.

Experiments were conducted on three bioinformatics graph benchmark datasets from \cite{morris2020tudataset}. The results demonstrate that the proposed refinement strategy improves the alignment between generated graphs and the statistical properties of the training data. The refined graphs show more coherent and realistic structural patterns, with connectivity that better reflects the relationships encoded in node features. Overall, the GA refinement reduces structural mismatches while maintaining variability, leading to higher-quality synthetic graphs suitable for downstream graph learning tasks.

\textbf{Data and Code:} These are both available at: \texttt{ github.com/shorinbonsai/WGAN-GA-Refine}.

\begin{figure}
    \centering 
    \includegraphics[width=1.0\linewidth]{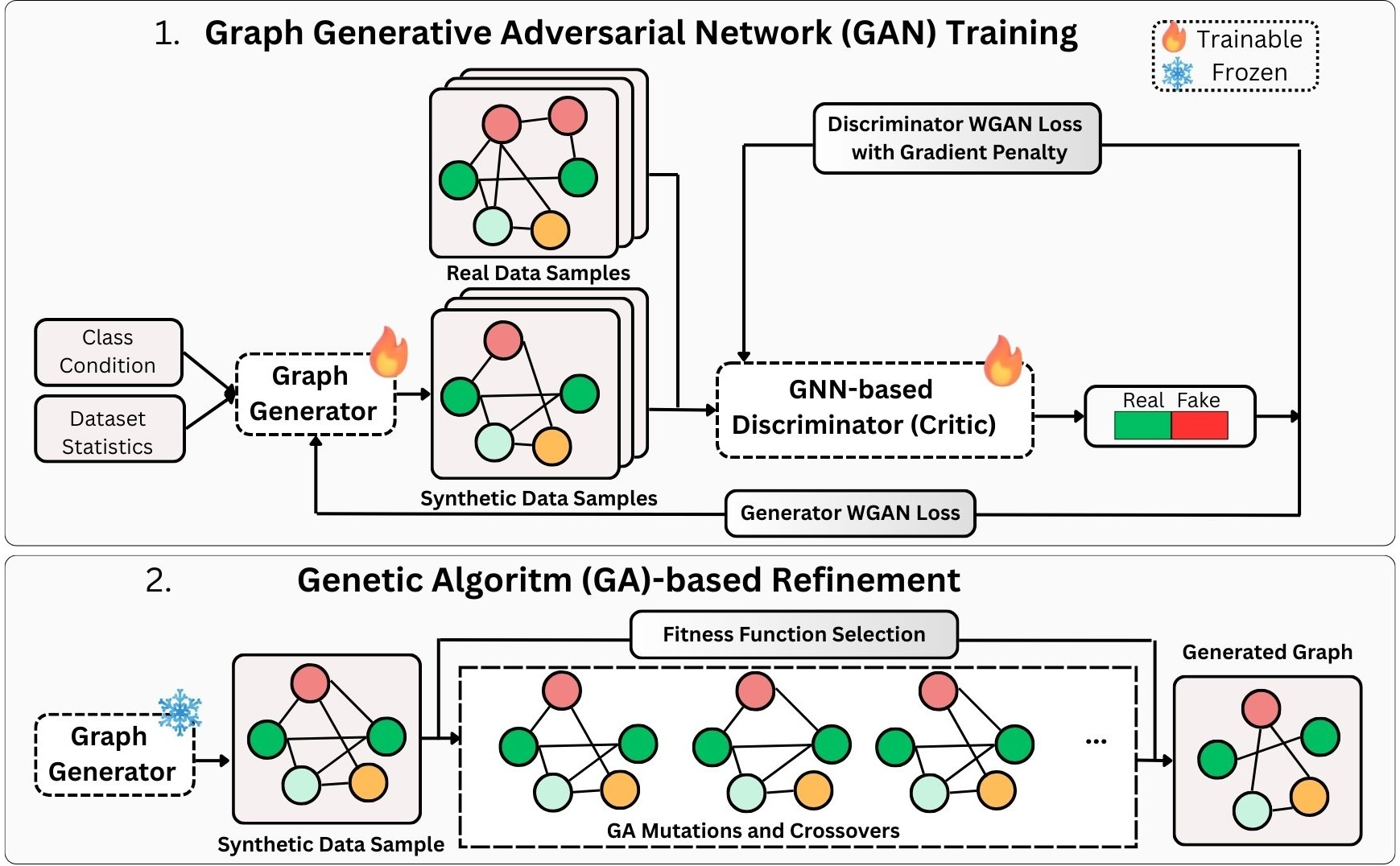}
    \caption{
Model overview. Phase 1: GAN training (based on~\cite{ava_cascon,ava_arxiv}). The generator learns to place nodes in a latent space where closer nodes are more likely to be connected. A GNN-based critic processes graphs using several convolution layers, pools node features, and combines them with class embeddings to compute Wasserstein scores. Phase 2: Genetic Algorithm (GA) refinement. After training, the generator produces synthetic graphs that are then refined using a GA that produces graphs closely resembling real samples. The fitness function encourages the generated graphs to match the real data distribution and target class while remaining distinct and novel.}
    \label{fig:arch}
\end{figure}

\section{Methodology: Evolutionary Refinement}
\label{sec:methodology}
Our framework uses the output of a deep generative model as input to evolutionary refinement to produce precise graph structures.
These stages are shown in Figure \ref{fig:arch}: (1) a coarse generation stage using a Wasserstein Generative Adversarial Network (WGAN) (based on~\cite{ava_cascon,ava_arxiv}) and (2) a refinement stage using a Genetic Algorithm (GA) library for edge editing developed in Rust.
We note that the focus of this paper is the evolutionary refinement stage, and that the coarse generation stage could use any method that produces graph structures. Although this method can be applied to extend existing GAN-based graph generation models, we use WGAN as a representative example due to its strong reported performance.

\subsection{Coarse Generation: WGAN}

WGAN generates coarse graph candidates that capture the global structural characteristics of the target graph distribution. Graphs are represented as adjacency matrices, with the generator mapping latent samples to continuous adjacency estimates. The critic is trained using the Wasserstein distance with a Lipschitz constraint, which provides a smoother and more stable training signal than standard GANs.

The primary role of the WGAN in our system is not to produce perfectly valid graphs, but to capture high-level statistical regularities observed in the training data, such as edge density, degree patterns, and broad spectral or clustering properties. These coarse outputs serve as informed initializations for the evolutionary stage. By starting the GA-based refinement from WGAN-generated candidates instead of random graphs, the search space explored during refinement is biased toward structurally plausible regions, improving convergence speed and solution quality.
As a result, the WGAN provides a global initialization of graph structure, while detailed refinement and structural optimization are handled by the evolutionary stage. The WGAN implementation is based on the code from~\cite{ava_arxiv}.

\subsection{Fine-Tuning: Evolutionary Refinement}

While GAN-based models capture global structural trends, they often struggle to satisfy precise local constraints, such as exact degree distributions or triangle counts. To address this limitation, we refine the generated graphs using a GA.

In this work, we provide a  Rust-based GA library that integrates WGAN-based graph generation with a GA refinement module. The Python environment interfaces with our library to pass graphs generated by the WGAN directly to the GA initialization stage. The GA then iteratively refines graph structures by optimizing a fitness function defined over selected structural metrics (e.g. clustering coefficient, spectral gap), that are also reflected in the WGAN training objective.

\subsection{Representation and Initialization}
\label{subsec:representation}

To effectively bridge the continuous latent space of the WGAN and the discrete search space of the graph refiner, we utilize a dual-representation strategy. This approach distinguishes between the \textit{phenotype} (the actual graph structure) and the \textit{genotype} (the evolutionary instruction set).
\subsubsection{Graph Phenotype}
The phenotype represents the physical graph structure utilized for fitness evaluation. It is implemented in the GA library as an adjacency list.

\subsubsection{Command String Genotype}

Our system employs a linear command-based genotype. Each individual in the population is defined by a genome $g$, a sequence of genes that each encode a specific operation. The final graph is produced by an expression function $\Phi$:
\begin{equation}
    G_{final} = \Phi(G_{base}, g) = op_n(\dots op_2(op_1(G_{base}))\dots)
\label{eqn:GA}
\end{equation}
where $G_{base}$ is the coarse graph generated by the WGAN. This ensures that the GA explores the local neighbourhood of the coarse graph, rather than the entire search space.
The evolutionary search progresses through the manipulation of the linear command genotype. 

\subsection{Genetic Operators and Dynamics}

Crucially, our operators do not directly manipulate edges on the adjacency matrix; instead, they manipulate the \textit{instruction sequence} that constructs the graph. 

In particular, the chromosome is a sequence of edge-editing operations that is applied sequentially; it is thus deterministic and specifies a particular network.
The length of the chromosome is twice the number of vertices, to stay relatively consistent with \cite{CEC19} while scaling for graphs with different cardinalities.

 \subsubsection{Edge Editing Operations}
\label{subsubsec:operations}
The operations are applied to a graph $G(V, E)$ with nodes $u$, $v$, $w$, and $x$ from $V$, as follows.
\begin{itemize}
    \item \textbf{Toggle($u$, $v$):} If edge $\{u, v\}$ is in $E$ then remove it from $E$, else add it to $E$. \textbf{\textit{Local} Toggle}($u$, $w$, $v$) requires that edges $\{u, w\}$ and $\{w, v\}$ exist then calls \textbf{Toggle($u$, $v$)}.
    \item \textbf{Hop($u$, $v$, $w$):} If edges $\{u, v\}$ and $\{v, w\}$ are in $E$ and edge $\{u, w\}$ is not, then remove $\{u, v\}$ from $E$ and add $\{u, w\}$ to $E$.
    \item \textbf{Add($u$, $v$):} If $\{u, v\}$ is not in $E$ then add it to $E$, else do nothing.
    \textbf{\textit{Local} Add($u$, $w$, $v$)} requires that edges $\{u, w\}$ and $\{w, v\}$ exist then calls \textbf{Add}($u$, $v$).
    \item \textbf{Delete($u$, $v$):} If $\{u, v\}$ is in $E$ then remove it from $E$, else do nothing. 
    \textbf{\textit{Local} Delete($u$, $w$, $v$)} requires that edges $\{u, w\}$ and $\{w, v\}$ exist then calls \textbf{Delete}($u$, $v$).
    \item \textbf{Swap($u$, $v$, $w$, $x$):} If $\{u, v\}$ and $\{w, x\}$ are the only edges in $E$ between nodes $u$, $v$, $w$ and $x$ then remove them from $E$ and add $\{u, x\}$ and $\{v, w\}$ to $E$.
    \item \textbf{Null():} Do nothing.
\end{itemize}

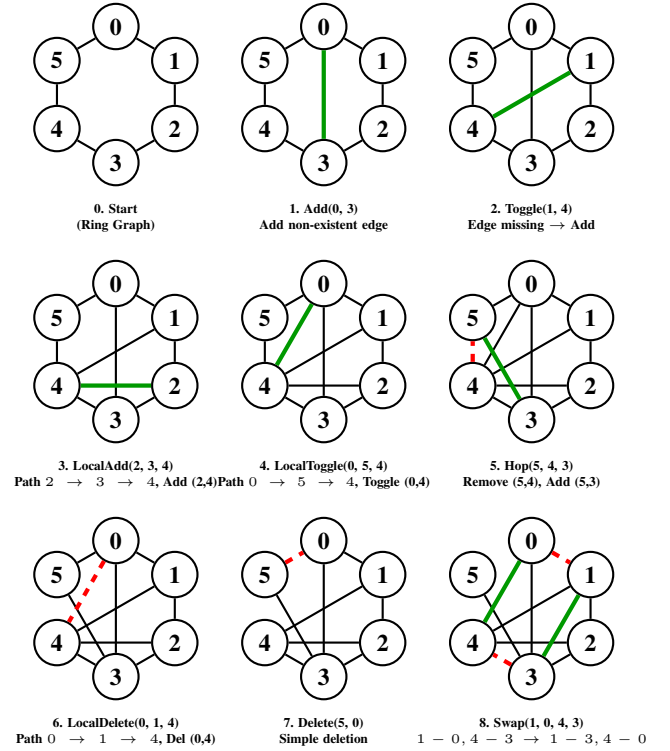
\begin{figure}[htbp]
    \centering

    \begin{tikzpicture}[x=2.75cm, y=3.4cm]

        \newcommand{\drawstate}[6]{
            \begin{scope}[shift={(#1,#2)}]
                \node[op_label] at (0, -1.4cm) {#3};

                \foreach \i in {0,...,5} {
                    \node[vertex] (v\i) at ({90 - \i*360/6}:0.9cm) {\i};
                }

                \foreach \u/\v in {#4} {
                    \draw[edge] (v\u) -- (v\v);
                }

                \foreach \u/\v in {#5} {
                    \draw[new_edge] (v\u) -- (v\v);
                }

                \foreach \u/\v in {#6} {
                    \draw[del_edge] (v\u) -- (v\v);
                }
            \end{scope}
        }

        \drawstate{0}{0}
            {0. Start\\(Ring Graph)}
            {0/1, 1/2, 2/3, 3/4, 4/5, 5/0} 
            {} 
            {}

        \drawstate{1}{0}
            {1. Add(0, 3)\\Add non-existent edge}
            {0/1, 1/2, 2/3, 3/4, 4/5, 5/0} 
            {0/3} 
            {}

        \drawstate{2}{0}
            {2. Toggle(1, 4)\\Edge missing $\to$ Add}
            {0/1, 1/2, 2/3, 3/4, 4/5, 5/0, 0/3} 
            {1/4} 
            {}

        \drawstate{0}{-1}
            {3. LocalAdd(2, 3, 4)\\Path $2\to3\to4$, Add (2,4)}
            {0/1, 1/2, 2/3, 3/4, 4/5, 5/0, 0/3, 1/4} 
            {2/4} 
            {}

        \drawstate{1}{-1}
            {4. LocalToggle(0, 5, 4)\\Path $0\to5\to4$, Toggle (0,4)}
            {0/1, 1/2, 2/3, 3/4, 4/5, 5/0, 0/3, 1/4, 2/4} 
            {0/4} 
            {}

        \drawstate{2}{-1}
            {5. Hop(5, 4, 3)\\Remove (5,4), Add (5,3)}
            {0/1, 1/2, 2/3, 3/4, 5/0, 0/3, 1/4, 2/4, 0/4} 
            {5/3} 
            {4/5} 

        \drawstate{0}{-2}
            {6. LocalDelete(0, 1, 4)\\Path $0\to1\to4$, Del (0,4)}
            {0/1, 1/2, 2/3, 3/4, 5/0, 0/3, 1/4, 2/4, 5/3} 
            {} 
            {0/4} 

        \drawstate{1}{-2}
            {7. Delete(5, 0)\\Simple deletion}
            {0/1, 1/2, 2/3, 3/4, 0/3, 1/4, 2/4, 5/3} 
            {} 
            {5/0} 

        \drawstate{2}{-2}
            {8. Swap(1, 0, 4, 3)\\$1-0, 4-3 \to 1-3, 4-0$}
            {1/2, 2/3, 0/3, 1/4, 2/4, 5/3} 
            {1/3, 0/4} 
            {0/1, 3/4}

    \end{tikzpicture}
    \caption{Sequential visualization of the edge editing operations. Green edges are added; red dashed edges are removed. The chromosome producing the final graph (bottom right) from the start graph (top left) thus consists of these 8 operations in sequence.}
    \label{fig:graph_ops}
\end{figure}

These are visualized in a sequential manner in Figure \ref{fig:graph_ops}.
The probabilities of the operations are given in Table \ref{tab:GAsettings}. In initial tuning, the probabilities of Toggle, Add, Delete, and Local toggle were all set to half those of the other operations because otherwise the degree distribution was negatively affected. 

\subsubsection{Crossover}
We employ Two-Point Crossover to recombine successful edit strategies from the population. 

This method is preferred over Uniform Crossover for this application as it preserves contiguous subsequences of graph operations. Since the phenotype is constructed by applying commands sequentially, preserving these functional blocks allows offspring to inherit effective macro-actions (e.g., a specific sequence of edits that creates a valid substructure) without disrupting the logic of the generative process.

\subsubsection{Mutation}
Mutation randomly selects 1--4 genes and replaces those with new random genes.
This slightly disruptive operation promotes a deeper exploration of the search space.

\subsection{Fitness Function}
\label{subsec:fitness}
The fitness function quantifies the structural deviation between a candidate graph $G$ and the distribution of the target graphs.

Since the goal is to generate graphs that statistically resemble a class of real-world networks rather than reproducing a single instance, we employ MMD.

To compute the final score, we treat the features of the generated graph as a sample and compare them to the distribution of features in the training set. The total fitness $F$ (where lower is better) is calculated as a weighted sum of the MMD scores combined with a penalty for deviation in graph density:
\begin{equation}
    F = (w_d \cdot \text{MMD}_d) + (w_c \cdot \text{MMD}_c) + (w_s \cdot \text{MMD}_s) + P_{\text{edge}}
\end{equation}
where $w_d, w_c,$ and $w_s$ are configurable weights for the degree, clustering, and spectral components respectively, chosen to be consistent with \cite{ava_arxiv}. The edge penalty term $P_{\text{edge}}$ ensures the graph maintains a realistic density and is defined as

$P_{\text{edge}} = w_e \cdot | |E_{gen}| - E_{target} |$,
where $|E_{gen}|$ is the current edge count, $E_{target}$ is the desired edge count, and $w_e$ is the penalty weight.

 The GA minimizes this aggregate distance, driving the population toward graphs that are statistically authentic in both local detail and global organization.
 The components of the fitness landscape are as follows.

\subsubsection{Degree Distribution ($\text{MMD}_d$)}
To ensure the generated graphs reproduce realistic connectivity patterns, we compute the degree distribution histogram for the candidate graph. Let $H_d(G)$ be the normalized histogram of node degrees for graph $G$, computed using adaptive binning with a static count of ten bins to accommodate varying graph sizes. $\text{MMD}_d$ measures the distance between $H_d(G)$ and the set of degree histograms. This penalizes graphs that fail to capture the density and distribution of the real data.

\subsubsection{Clustering Coefficient ($\text{MMD}_c$)}
Local substructures are evaluated using the distribution of local clustering coefficients. For each node $v$, the clustering coefficient $C_v$ measures the density of triangles in its neighbourhood. $\text{MMD}_c$ measures the distance between the graph and the data distribution. This metric is crucial for enforcing transitivity and community structure, properties often missed by simpler models.

\subsubsection{Spectral Features ($\text{MMD}_s$)}
Global structural properties are captured using the graph spectrum. We compute the eigenvalues of the combinatorial Laplacian matrix 
$\mathbf{L} = \mathbf{D}-\mathbf{A}$, where $\mathbf{A}$ is the adjacency matrix and $\mathbf{D}$ is the degree matrix. The eigenvalues are computed and sorted. This spectrum encodes vital information regarding graph cuts, connectivity, and diffusion properties. By minimizing the MMD between the spectral densities, the GA ensures that the generated graphs share the same global topology as the target class.

\section{Experimental Design}
\label{sec:experiments}

\subsection{Datasets}
We assess our graph refinement methods using three benchmark graph classification datasets~\cite{morris2020tudataset} that cover diverse domains and structural complexities, detailed in Table \ref{tab:dataset_characteristics} below. The MUTAG dataset consists of chemical compound graphs for mutagenicity prediction and is characterized by relatively small graphs with an average of 18 nodes.
ENZYMES represents protein structures distributed across six classes, consisting of moderate-sized graphs averaging 33 nodes.
Finally, PROTEINS serves as the largest and most structurally varied dataset for distinguishing enzymes from non-enzymes, with graph sizes spanning from 4 to 620 nodes.
The feature representations differ significantly among these benchmarks: while MUTAG incorporates both 7-dimensional node and 4-dimensional edge features, ENZYMES and PROTEINS rely exclusively on 3-dimensional node features. Collectively, these datasets provide a rigorous environment for testing graph generation performance across varying scales and complexities.

\begin{table}[]
\centering
\begin{center}
\caption{Summary of Dataset Characteristics}
\begin{tabular}{lccc}
\toprule
\textbf{Property} & \textbf{MUTAG} & \textbf{ENZYMES} & \textbf{PROTEINS} \\
\midrule
\textbf{Domain} & Chemistry & Biochemistry & Biochemistry \\
\textbf{Number of Graphs} & 188 & 600 & 1,113 \\
\textbf{Number of Classes} & 2 & 6 & 2 \\
\textbf{Avg. Nodes} & 17.93 & 32.63 & 39.06 \\
\textbf{Avg. Edges} & 19.79 & 62.14 & 72.82 \\

\bottomrule
\end{tabular}
\end{center}
\label{tab:dataset_characteristics}
\end{table}

\begin{table}[]
    \centering
    \caption{Settings for the Evolutionary Algorithm}
    \begin{tabular}{|c|c|}
    \hline
    Parameter     &  Settings \\ \hline \hline
    Crossover/Mutation     & 50/80 \\ \hline
    Tournament Size        & 5 \\ \hline
    Population Size        & 500 \\ \hline
    Generations            & 300 \\ \hline
    Elitism                & 2 \\ \hline
    Operation Probability    &   
        \begin{tabular}{@{}l@{: }l l@{: }l l@{: }l@{}}
            Toggle & 1/14 & Delete & 1/14 & LocalAdd & 1/7 \\
            Hop & 1/7 & Swap & 1/7 & LocalDel & 1/7 \\
            Add & 1/14 & LocalTog & 1/14 & Null & 1/7 \\
        \end{tabular} \\ \hline
    \end{tabular}
    \label{tab:GAsettings}
\end{table}

\begin{table*}[]
\centering
\caption{Comparison of WGAN vs. GA-Refined Graphs. Lower MMD scores indicate better distributional matching. }
\label{tab:refined_comparison}
\resizebox{\textwidth}{!}{%
\begin{tabular}{ll c cc cc cc cc}
\toprule

\multirow{2}{*}{\textbf{Dataset}} & \multirow{2}{*}{\textbf{Class}} & \textbf{Nodes} & \multicolumn{2}{c}{\textbf{Avg Edges}} & \multicolumn{2}{c}{\textbf{MMD Degree}} & \multicolumn{2}{c}{\textbf{MMD Clustering}} & \multicolumn{2}{c}{\textbf{MMD Spectral}} \\
\cmidrule(lr){4-5} \cmidrule(lr){6-7} \cmidrule(lr){8-9} \cmidrule(lr){10-11}

& & (Real$\rightarrow$WGAN) & WGAN & \textbf{Refined} & WGAN & \textbf{Refined} & WGAN & \textbf{Refined} & WGAN & \textbf{Refined} \\
\midrule

\multirow{2}{*}{\textbf{MUTAG}} 
 & Class 0 & 13.4$\rightarrow$13.8 & 14.7 & 14.0 & 0.259 & \textbf{0.247} & 1.070 & \textbf{0.000} & \textbf{0.081} & 0.112 \\
 & Class 1 & 20.7$\rightarrow$18.6 & 19.7 & 21.5 & 0.295 & \textbf{0.270} & 1.043 & \textbf{0.000} & \textbf{0.113} & 0.130 \\
\midrule

\multirow{6}{*}{\textbf{ENZYMES}} 
 & Class 0 & 31.6$\rightarrow$38.4 & 76.4 & 66.6 & \textbf{0.210} & 0.213 & 0.204 & \textbf{0.200} & 0.198 & \textbf{0.034} \\
 & Class 1 & 32.1$\rightarrow$29.1 & 59.3 & 56.7 & \textbf{0.137} & 0.175 & \textbf{0.127} & 0.131 & 0.114 & \textbf{0.099} \\
 & Class 2 & 30.1$\rightarrow$30.5 & 60.3 & 56.0 & \textbf{0.156} & 0.165 & 0.139 & \textbf{0.135} & 0.053 & \textbf{0.034} \\
 & Class 3 & 37.4$\rightarrow$37.1 & 76.5 & 74.9 & \textbf{0.210} & 0.219 & \textbf{0.168} & \textbf{0.168} & 0.098 & \textbf{0.021} \\
 & Class 4 & 32.1$\rightarrow$28.4 & 54.4 & 58.6 & \textbf{0.147} & 0.152 & \textbf{0.115} & 0.116 & 0.053 & \textbf{0.048} \\
 & Class 5 & 38.8$\rightarrow$27.7 & 51.5 & 52.5 & \textbf{0.146} & 0.165 & 0.124 & \textbf{0.122} & \textbf{0.056} & 0.063 \\
\midrule

\multirow{2}{*}{\textbf{PROTEINS}} 
 & Class 0 & 54.5$\rightarrow$47.2 & 111.2 & 95.0 & \textbf{0.059} & 0.074 & 0.047 & \textbf{0.046} & 0.176 & \textbf{0.026} \\
 & Class 1 & 19.6$\rightarrow$24.0 & 52.0 & 40.1 & \textbf{0.091} & 0.110 & \textbf{0.057} & 0.062 & 0.076 & \textbf{0.071} \\

\bottomrule
\end{tabular}
}
\end{table*}

\subsection{GA Setup}
\label{subsec:gasetup}
The hyperparameters used for the evolutionary refinement stage are detailed in Table~\ref{tab:GAsettings}. The evolutionary process is managed by a Rust backend, which handles population management, fitness evaluation, and genetic operators in parallel.

\subsubsection{Population Initialization and the Identity Genome}
A critical component of our hybrid architecture is the initialization strategy. Standard evolutionary approaches often initialize populations with random noise; however, our goal is to \textit{refine} the structural priors learned by the WGAN, not to generate graphs from scratch. Consequently, the initialization process in the GA is biased toward the WGAN output.
The population of size $P$ is initialized as follows. First, the \textit{Identity Genome}, the first individual in the population, is explicitly constructed as a sequence consisting entirely of \textbf{Null}() operations; hence no changes are made to the base graph provided by the WGAN. This guarantees that the evolutionary search begins with a solution that is at least as fit as the original WGAN output, providing a stable baseline for improvement.
The remaining $P-1$ individuals are each initialized as a random sequence of operations, chosen according to the specified probabilities. This injects immediate diversity into the population, exploring the local structural neighbourhood of the base graph.

\subsubsection{Selection and Variation}
We employ tournament selection with a tournament size of 5. This balances exploration with exploitation of high fitness structural edits. To ensure that high quality refinements are never lost due to stochastic effects, we utilize elitism, preserving the top 2 individuals unchanged into the next generation.

We utilize high crossover (0.5) and mutation (0.8) probabilities. This aggressive variation strategy is viable because the operations are applied sequentially to a robust base structure, allowing the algorithm to rapidly test various combinations of edge additions, deletions, and swaps to correct specific spectral deviations and better exploit the search space.

\section{Results and Discussion}
\label{sec:results}

\subsection{Performance on Benchmark Datasets}

Table~\ref{tab:refined_comparison} presents a comprehensive study comparing the WGAN generator output against the final GA refined graphs. The results demonstrate that the evolutionary refinement of the base graphs significantly enhances the structural character, particularly capturing global topology and clustering patterns.

For the MUTAG dataset, the refinement yielded a dramatic improvement in the Clustering MMD, dropping from values greater than 1.0 in the base WGAN to near zero for both classes. This indicates that the GA successfully recovered the specific structures of the generally small graphs in the dataset that the WGAN struggled to capture. While the Spectral MMD increased slightly, the gains in clustering result and improved Degree MMD suggest more plausible synthetic graphs.

With the ENZYMES dataset, we see consistent and sometimes substantial improvements in the Spectral MMD, notably in Class 0 (0.198 to 0.034) and Class 3 (0.098 to 0.021). This suggests that the edge editing operations were successful in adjusting the graph eigenvalues to match the real distribution. In some of these cases we see a slight trade off, particularly with regards to Degree MMD, as the algorithm prioritized global connectivity over the degree counts.

The PROTEINS dataset further highlights the strength of the refinement in fixing topological properties. Spectral MMD improved significantly for both classes (e.g., Class 0: 0.176 to 0.026), and Clustering MMD remained competitive. The average edge counts in the refined graphs (95.0 for Class 0) dropped compared to the WGAN output (111.2), with the real data distribution for Class 0 being 103.

\subsection{Comparison to State of the Art}

\begin{table}[t!]
\centering
\caption{Comparison with state-of-the-art methods using MMD metrics (lower is better) as presented in~\cite{fan2019labeled,ava_cascon,ava_arxiv}}
\label{tab:mmd_comparison}
\begin{tabular}{lcccccc}
\toprule

 \textbf{Model} & \multicolumn{3}{c}{\textbf{PROTEINS}} & \multicolumn{3}{c}{\textbf{ENZYMES}} \\
 & Deg. & Clust. & Spect. & Deg. & Clust. & Spect.\\
\midrule
DeepGMG~\cite{li2018learning} & 0.96 & 0.63 & -- & 0.43 & 0.38 & -- \\
GraphRNN~\cite{you2018graph} & 0.04 & 0.18 & -- & 0.06 & 0.20 & -- \\
LGGAN~\cite{fan2019labeled} & 0.18 & 0.15 & -- & 0.09 & 0.17 & -- \\
WPGAN~\cite{ava_cascon} & \textbf{0.03} & 0.31 & -- & \textbf{0.02} & 0.28 & -- \\
Edge-WGAN~\cite{ava_arxiv} & 0.09 & 0.07 & 0.07 & 0.10 & 0.08 & 0.05  \\ 
\midrule
\textbf{(Ours) Refined} & 0.07 & \textbf{0.03} & \textbf{0.01} & 0.07 & \textbf{0.04} & \textbf{0.02}  \\
\bottomrule
\end{tabular}
\end{table}

Table~\ref{tab:mmd_comparison} compares the results of our refinement process against DeepGMG, GraphRNN, LGGA, and the base WGAN. The refined graphs outperform the baseline WGAN in Spectral MMD for both PROTEINS (0.01) and ENZYMES (0.02). 
In terms of Clustering MMD, the refined graphs achieve the best performance on PROTEINS (0.03) and remains highly competitive on ENZYMES (0.04), outperforming GraphRNN and LGGAN. While the degree distribution MMD is slightly higher than the best baseline in some cases, the overall balance of metrics indicates that the hybrid WGAN-GA approach generated graphs that are structurally more robust and representative of the target domain.

\subsection{Visual Analysis}

\begin{figure*}[t!]
    \centering
    \begin{minipage}{0.24\textwidth}
        \centering
        \includegraphics[width=\linewidth]{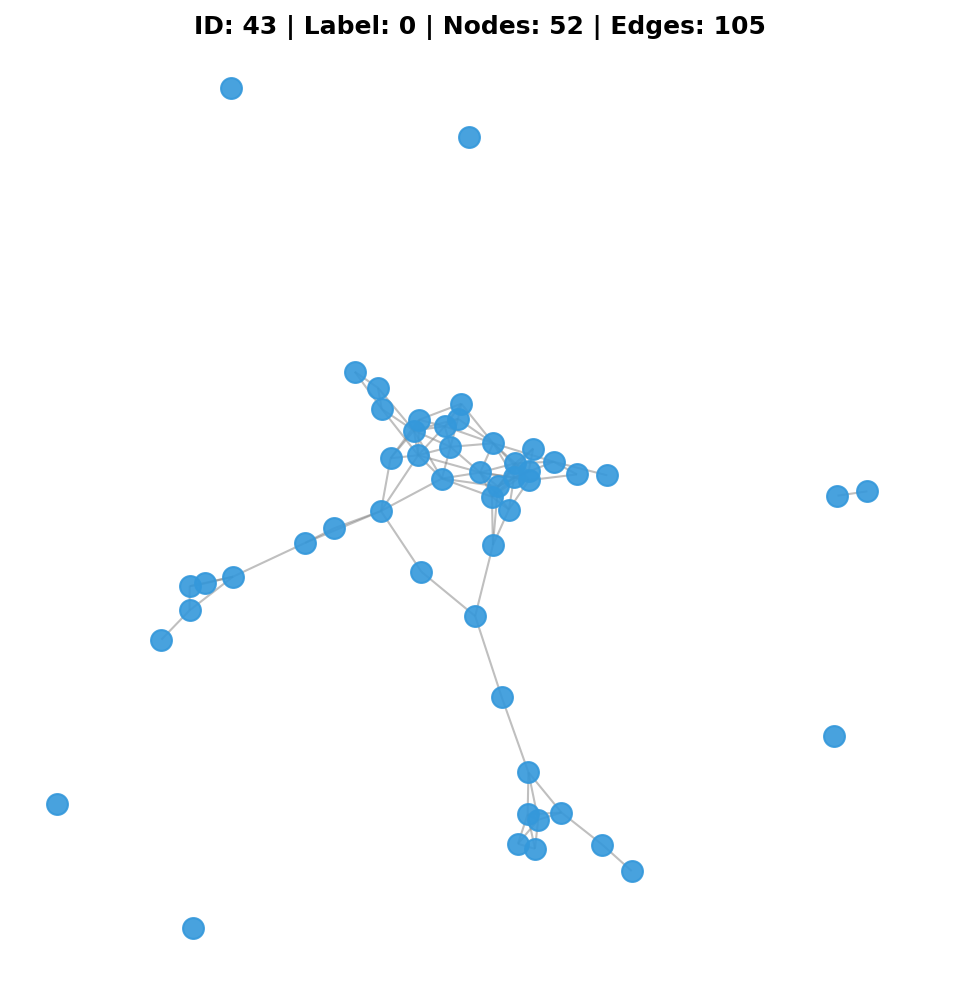}
    \end{minipage}\hfill
    \begin{minipage}{0.24\textwidth}
        \centering
        \includegraphics[width=\linewidth]{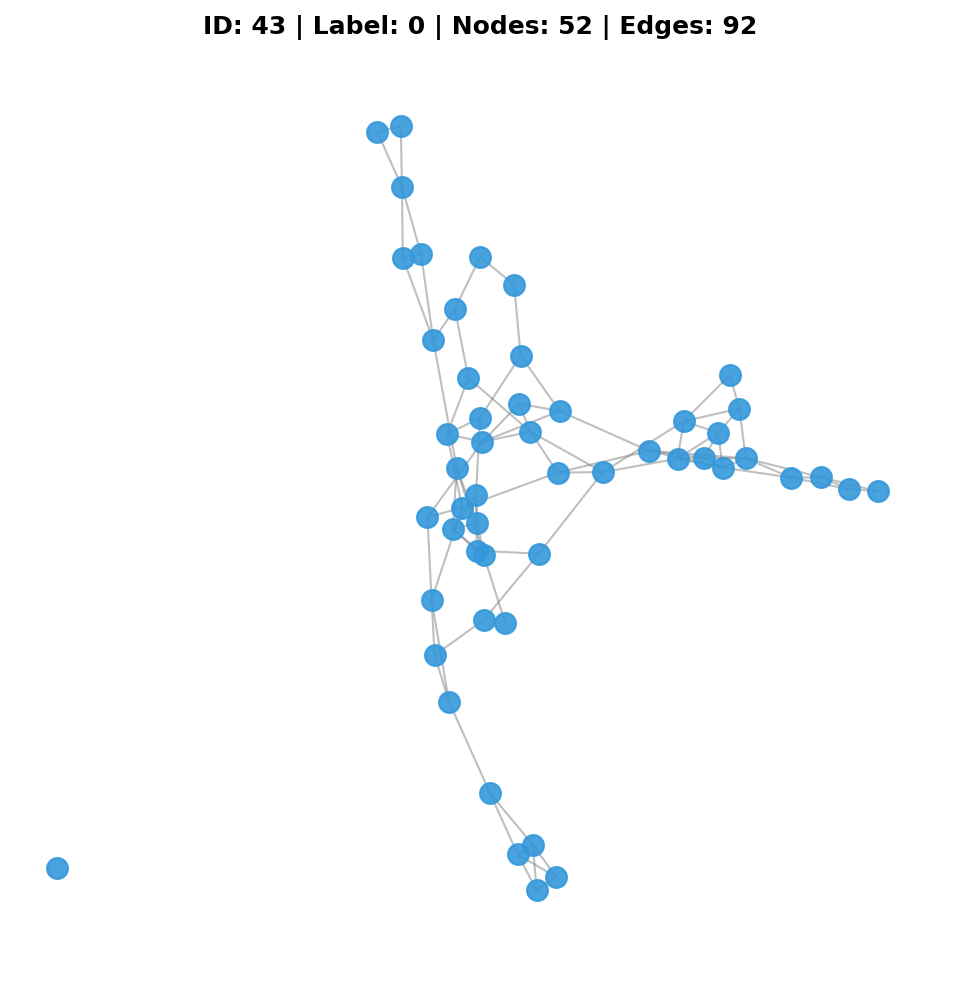}
    \end{minipage}
    \begin{minipage}{0.24\textwidth}
        \centering
        \includegraphics[width=\linewidth]{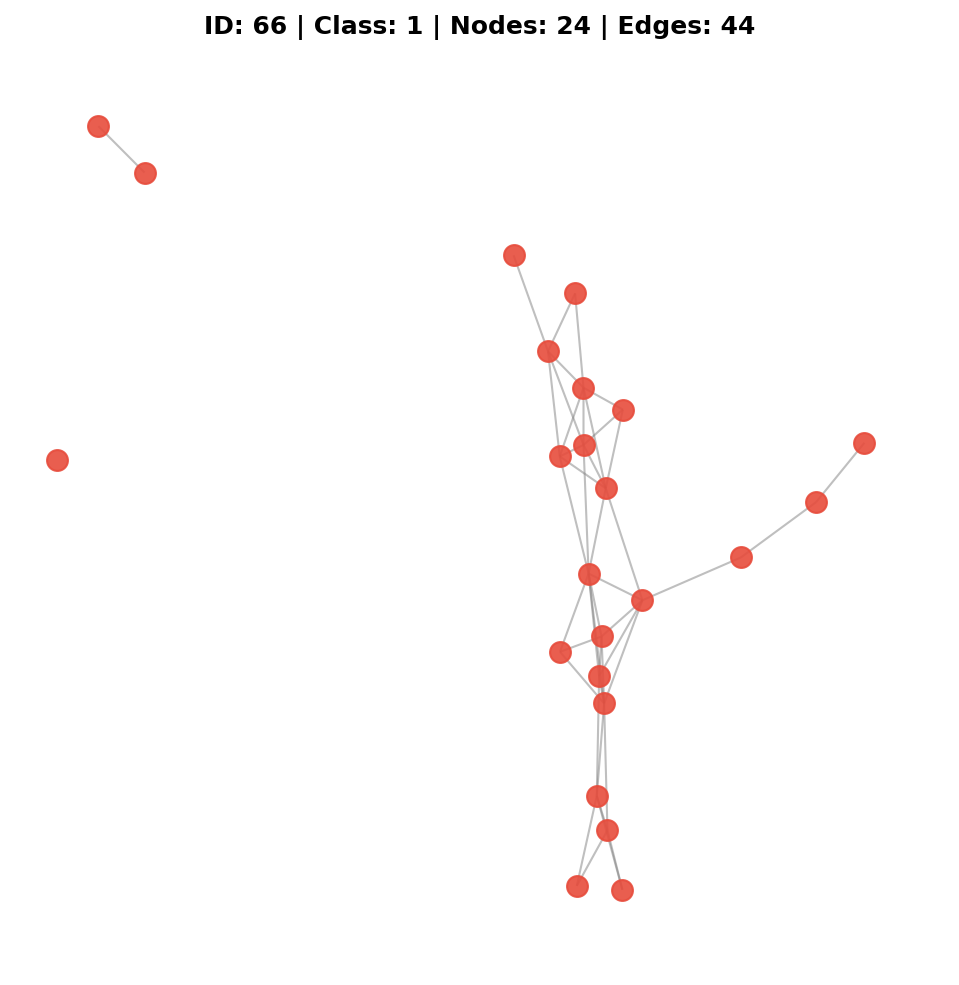}
    \end{minipage}\hfill
    \begin{minipage}{0.24\textwidth}
        \centering
        \includegraphics[width=\linewidth]{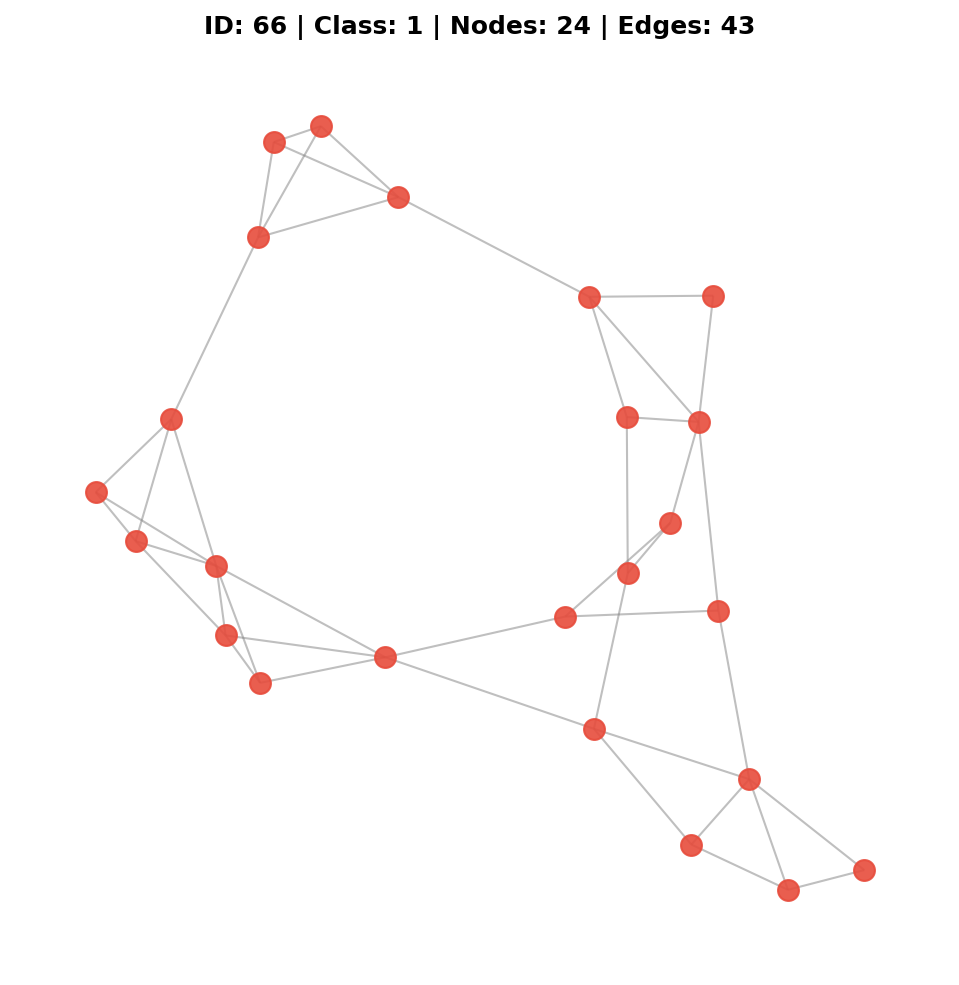}
    \end{minipage}\hfill
    
    \caption{Visual comparison on PROTEINS dataset (left: generated and refined graphs from Class 0; right: Class 1). }
    \label{fig:protein_vis}
\end{figure*}

\begin{figure*}[!htpb]
\centering
\begin{tabular}{ccc}
\includegraphics[width=0.3\textwidth]{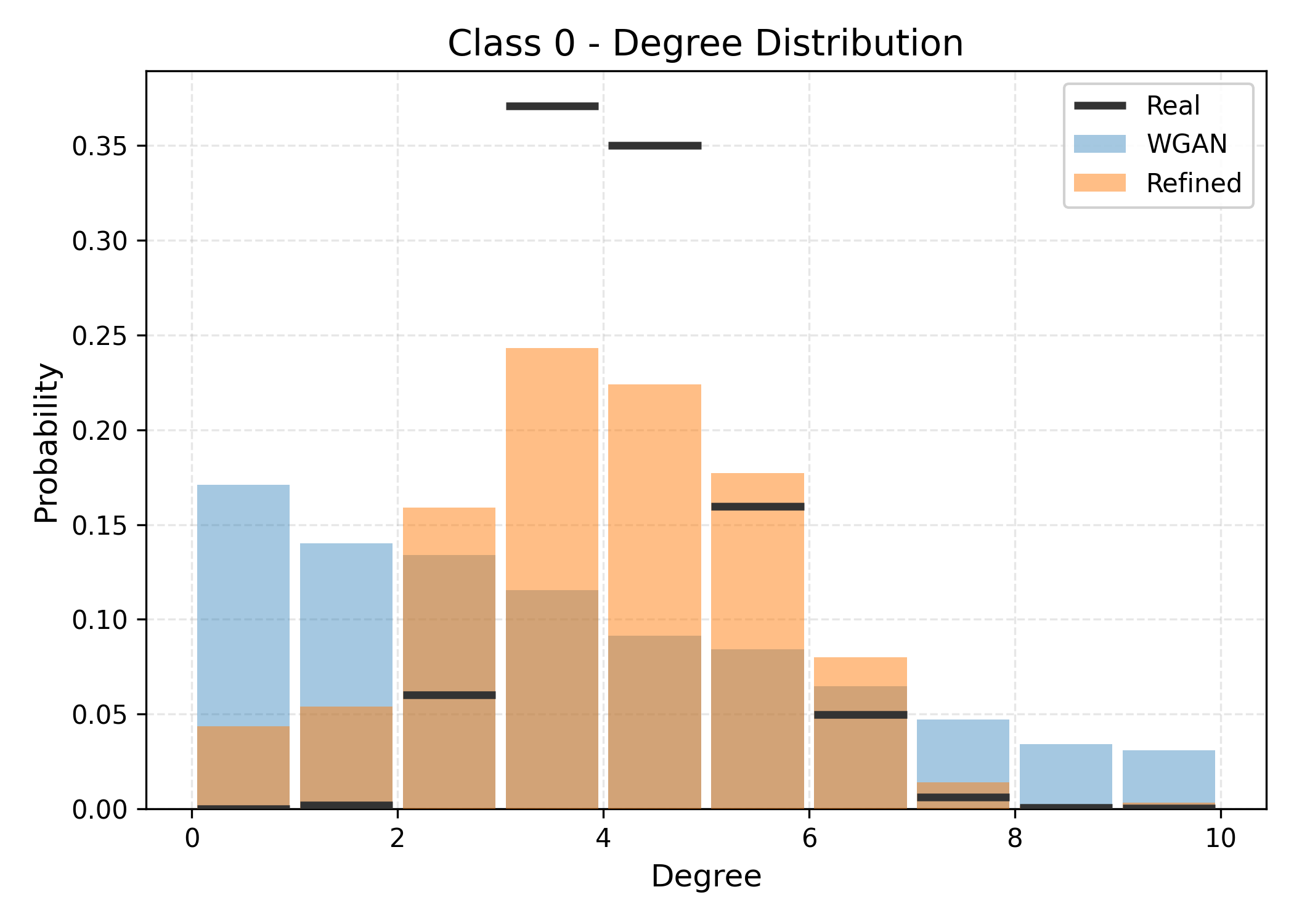}&
\includegraphics[width=0.3\textwidth]{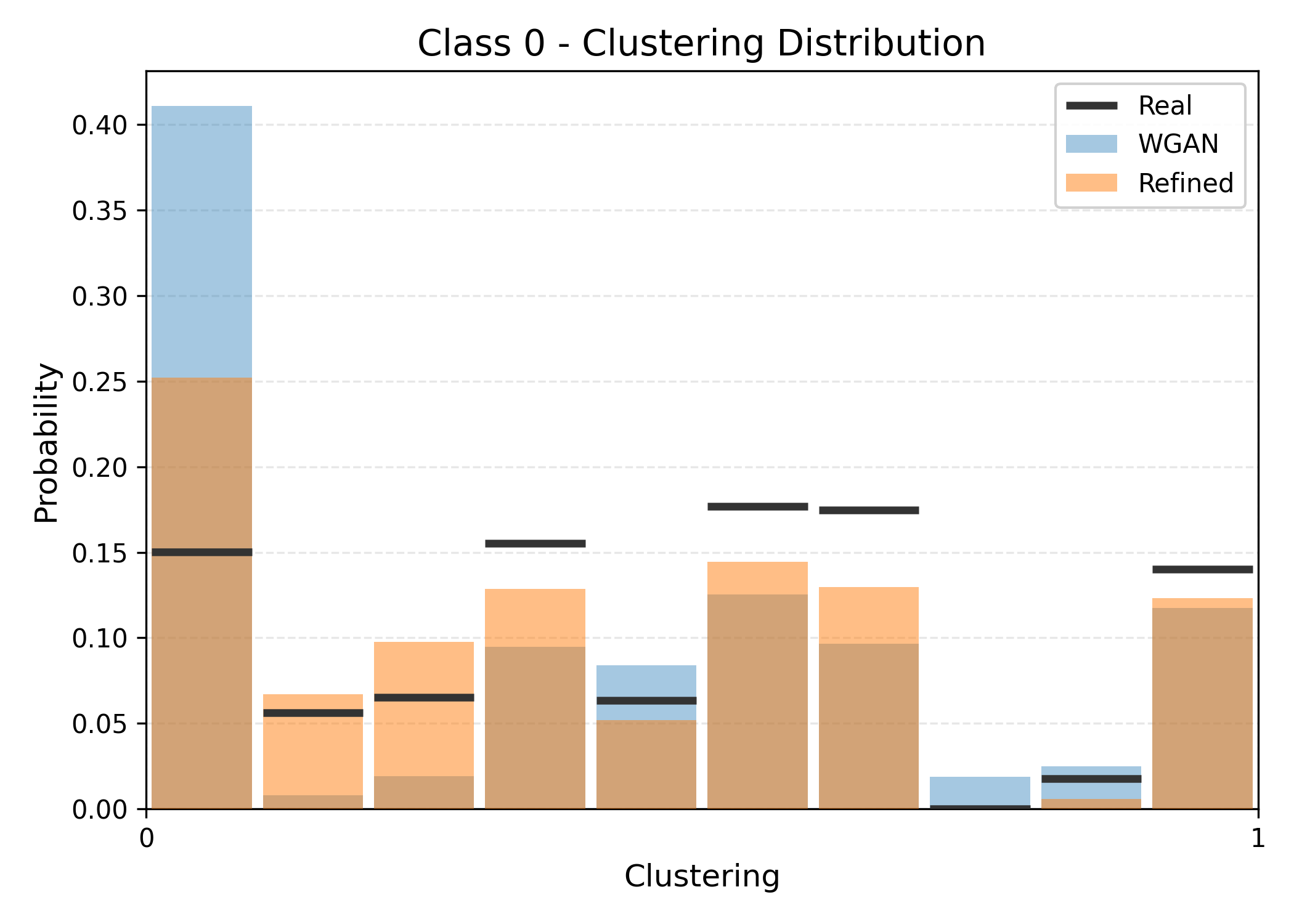}&
\includegraphics[width=0.3\textwidth]{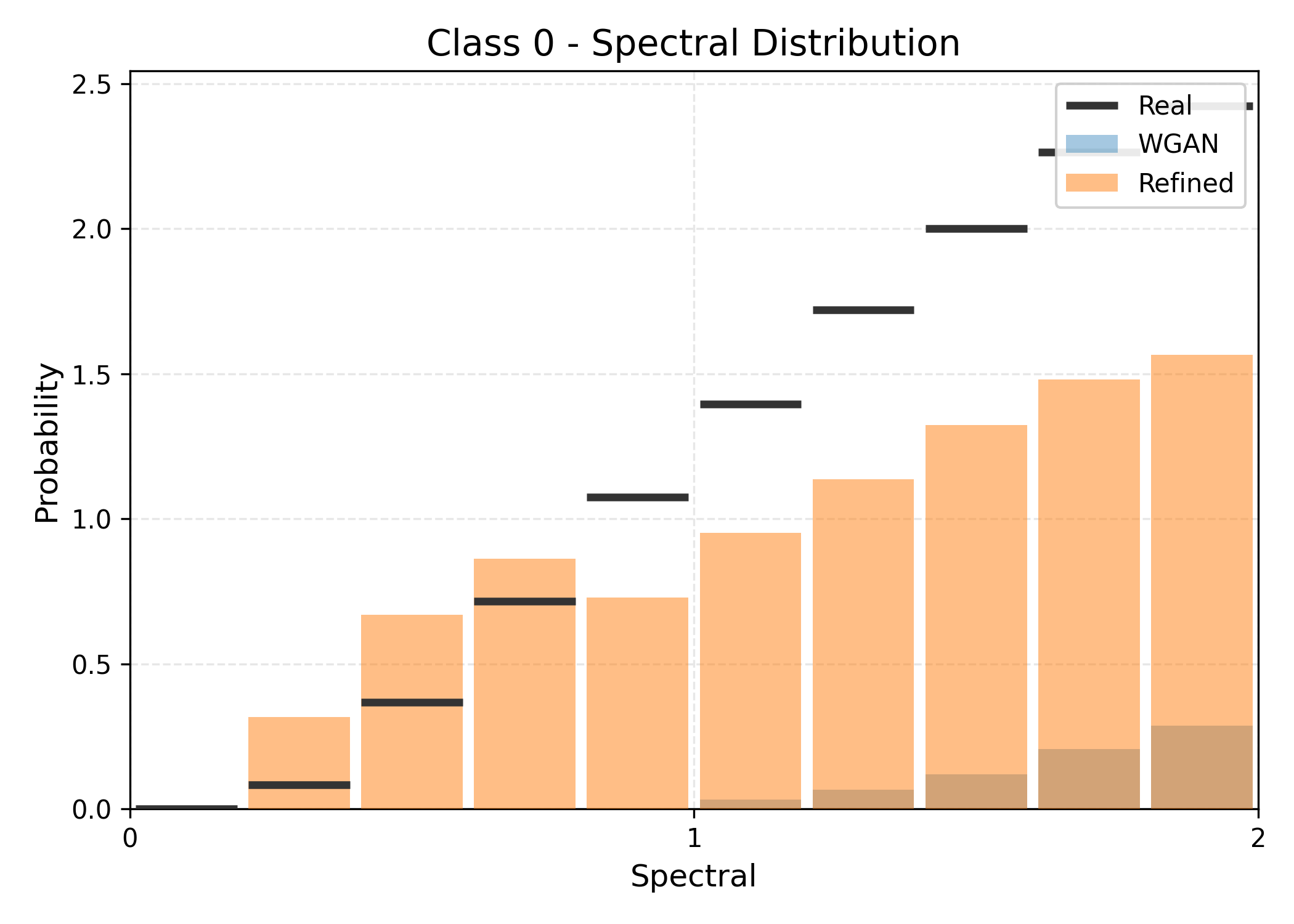}\\
\includegraphics[width=0.3\textwidth]{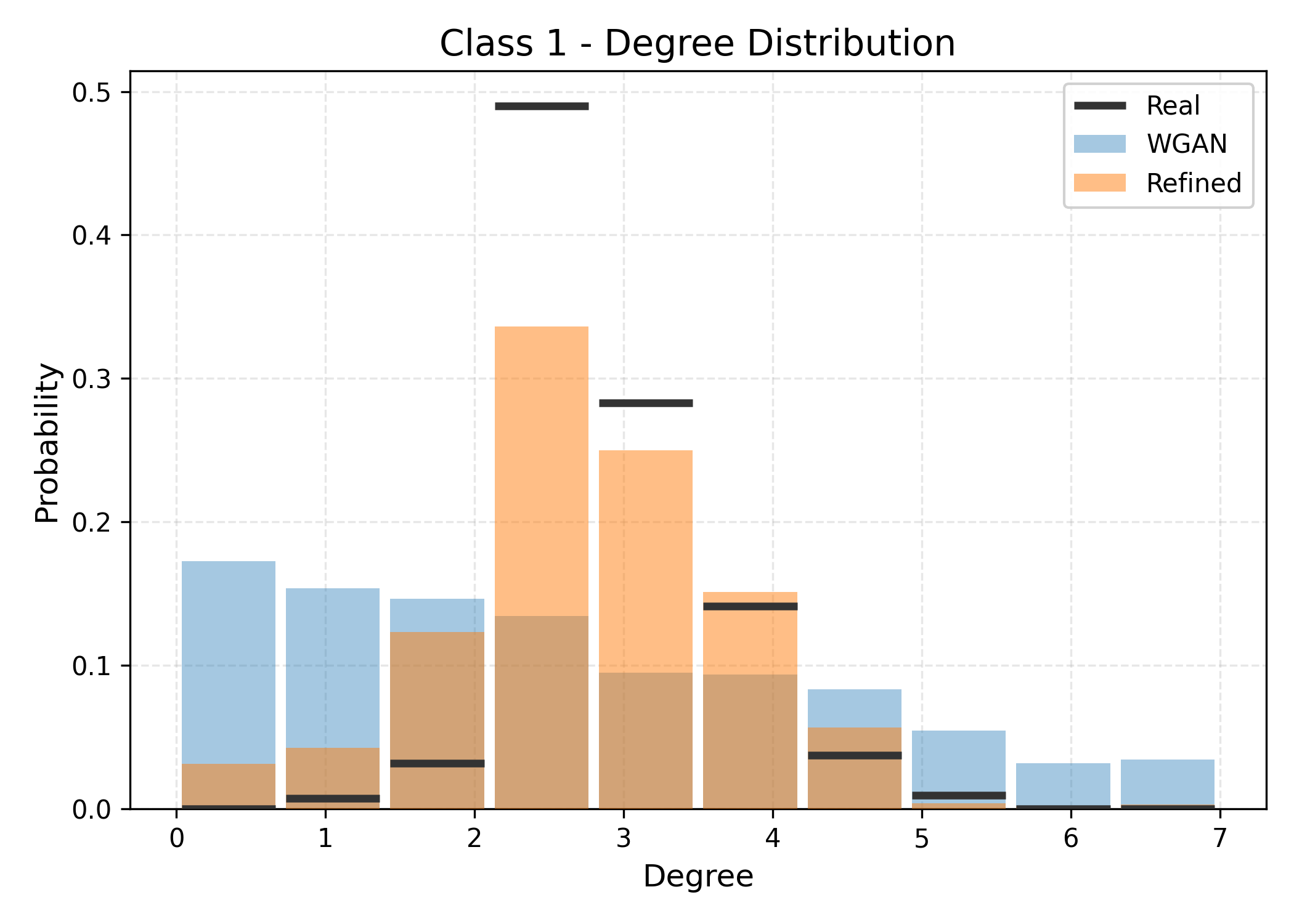}&
\includegraphics[width=0.3\textwidth]{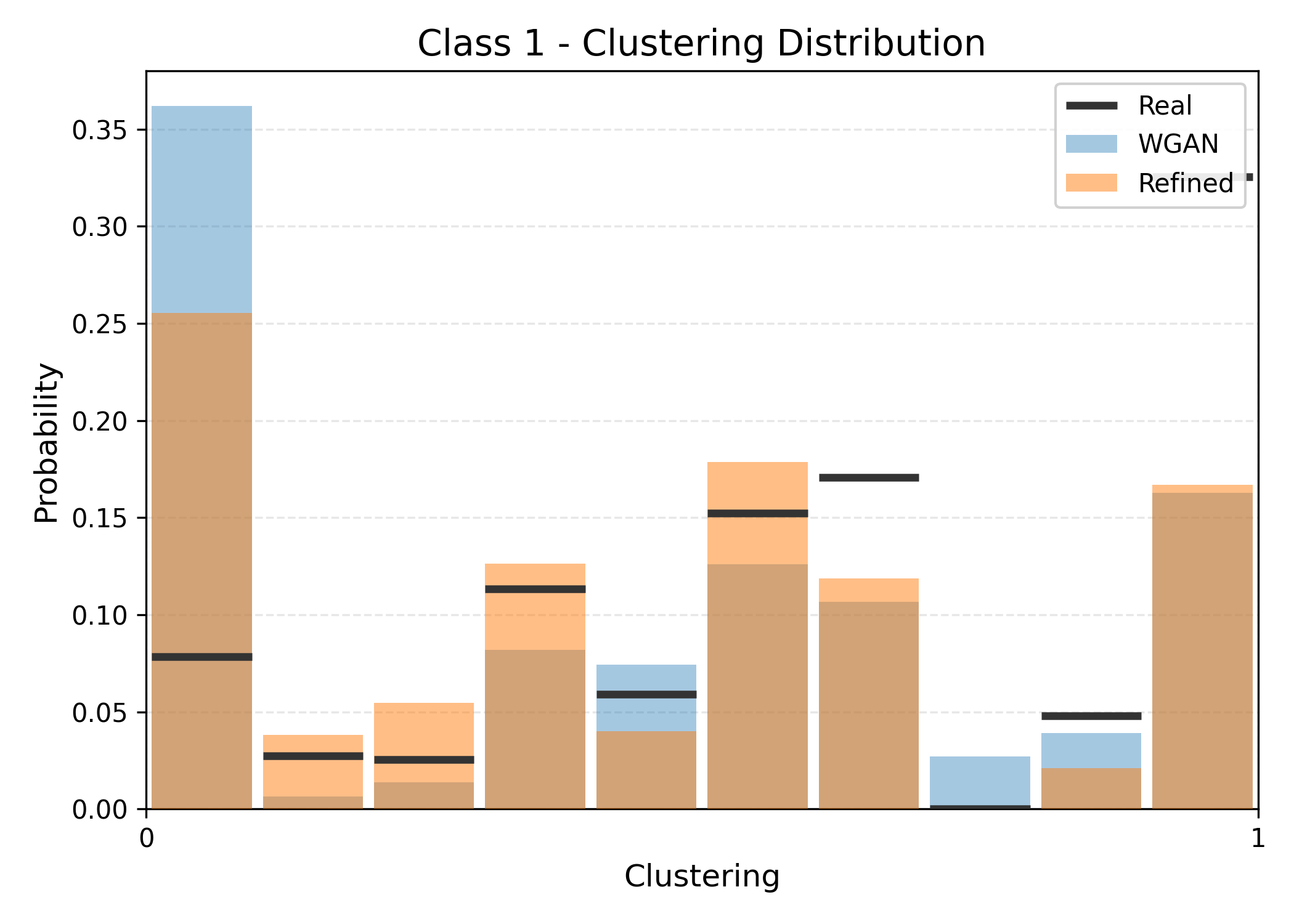}&
\includegraphics[width=0.3\textwidth]{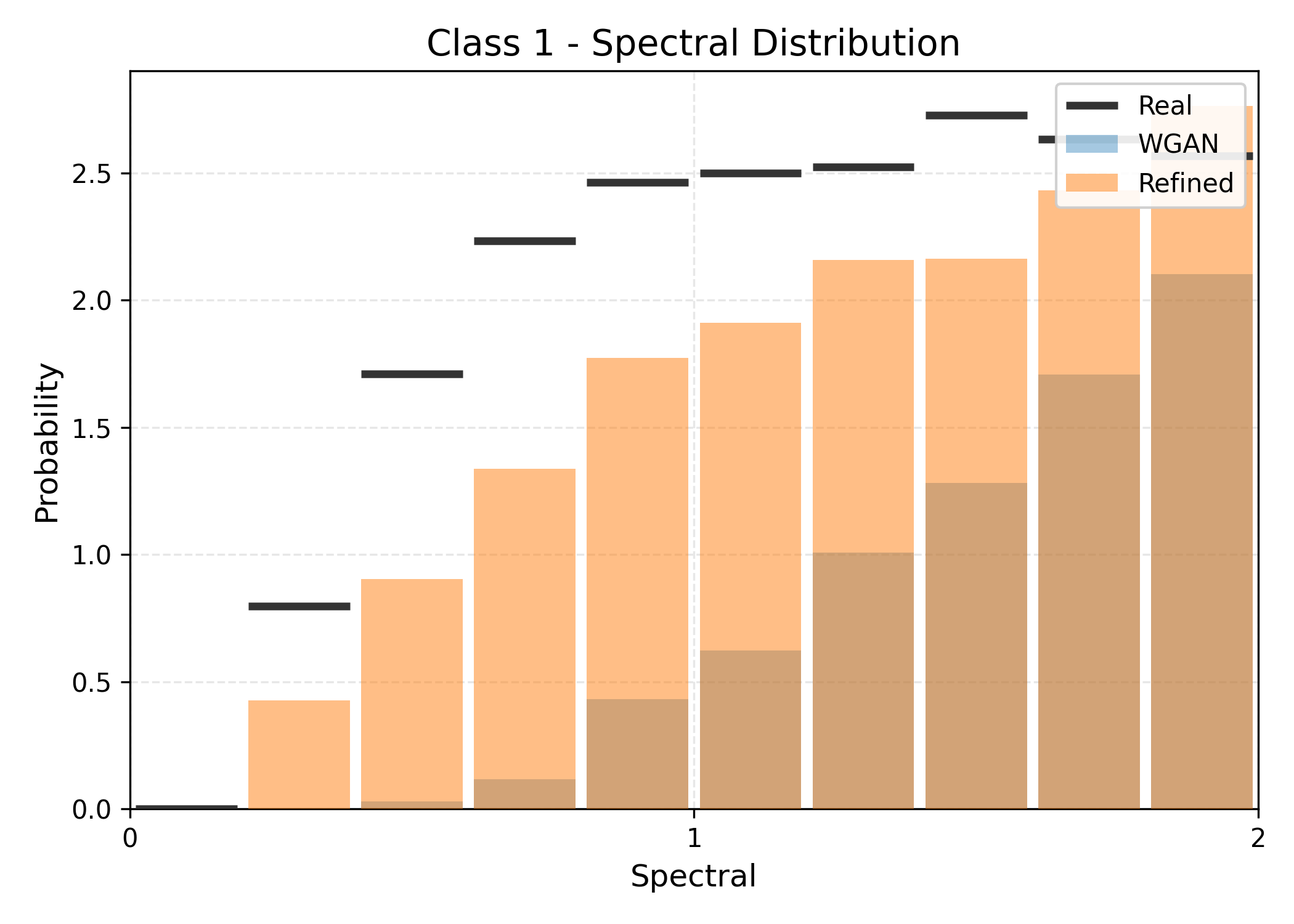}\\
\end{tabular}
\caption{Distribution comparison between real and generated graphs on the PROTEINS dataset.}
\label{fig:distribution}
\end{figure*}

Figure~\ref{fig:protein_vis} provides a qualitative comparison of the generated graphs. These two graphs were chosen as representative as they are both close to the mean cardinality of other graphs of their class. The refined graphs exhibit more coherent structures with fewer disconnected components compared to the raw WGAN outputs. This visual inspection correlates with the quantitative improvements in Spectral MMD, as the GA successfully connects isolated nodes and breaks unrealistic hubs not found in the source data. This connectivity also occurs while simultaneously decreasing the edge count in some cases to better match the data.

Figure~\ref{fig:distribution} displays the histograms for degree, clustering coefficient, and spectral distributions respectively. The refined distributions (orange) generally track the real data (black lines) more closely than the WGAN outputs (blue). Notably, the spectral distribution in Figure~\ref{fig:distribution} shows that the refinement process effectively shifts the eigenvalue spectrum to align much closer to the ground truth, correcting the skew in distribution seen in the coarser WGAN generation.

\section{Conclusions and Future Work}
\label{sec:conclusion}

In this work, we presented a framework for graph generation and refinement that combines global modelling of Wasserstein GANs with local optimization of GAs. By using the state of the art WGAN outputs as high quality initial seed populations, our GA is able to effectively refine the graph structure, correcting deviations in spectral and clustering that the WGAN on its own could not capture.

Our results on multiple benchmark datasets demonstrate that this two-stage approach consistently improves structural fidelity of the graphs that were already optimized using a GAN in \cite{ava_arxiv}. 
Specifically, the evolutionary refinement significantly lowers MMD scores for spectral and clustering distributions, giving us graphs that are topologically more realistic. Offloading this computationally heavy GA process to an optimized library ensures that the refinement is as efficient as possible. 

Future work could focus on expanding the ways in which the GA can be used in the graph generation, perhaps as an integral part of the generator process in graph based GANs. Researchers could also use the graph generation capability of the GA to replace the edge prediction component of the GAN, without it being stacked at the end of the training cycle.

\section*{Acknowledgements}

This work was supported in part by the Natural Sciences and Engineering Research Council of Canada (NSERC), and used the facilities of the Shared Hierarchical Academic Research Computing Network (SHARCNET) and Compute Canada.

\bibliography{reference}

\begin{thebibliography}{10}

\bibitem{arjovsky2017wasserstein}
M.~Arjovsky, S.~Chintala, and L.~Bottou.
\newblock Wasserstein generative adversarial networks.
\newblock In {\em Proc. Intl. Conf. on Machine Learning}, pages 214--223, 2017.

\bibitem{7529329}
A.~Barry, J.~Griffith, and C.~O'Riordan.
\newblock An evolutionary and graph-rewriting based approach to graph
  generation.
\newblock In {\em Intl. Joint Conf. on Computational Intelligence (IJCCI)},
  volume~1, pages 237--243, 2015.

\bibitem{de2018molgan}
N.~De~Cao and T.~Kipf.
\newblock {MolGAN: An implicit generative model for small molecular graphs}.
\newblock {\em ICML 2018 workshop on Theoretical Foundations and Applications
  of Deep Generative Models}, 2018.

\bibitem{CEC19}
M.~Dub\'{e}, S.~Houghten, and D.~Ashlock.
\newblock Representation for evolution of epidemic models.
\newblock In {\em 2019 IEEE Congress on Evolutionary Computation (CEC)}, pages
  2370--2377, Piscataway NJ, 06 2019. IEEE Press.

\bibitem{fan2019labeled}
S.~Fan and B.~Huang.
\newblock Labeled graph generative adversarial networks.
\newblock {\em arXiv preprint arXiv:1906.03220}, 2019.

\bibitem{gulrajani2017improved}
I.~Gulrajani et~al.
\newblock Improved training of {W}asserstein {GAN}s.
\newblock In {\em Advances in Neural Information Processing Systems}, 2017.

\bibitem{li2018learning}
Y.~Li et~al.
\newblock Learning deep generative models of graphs.
\newblock {\em arXiv preprint arXiv:1803.03324}, 2018.

\bibitem{morris2020tudataset}
C.~Morris et~al.
\newblock Tudataset: A collection of benchmark datasets for learning with
  graphs.
\newblock In {\em ICML 2020 Workshop on Graph Representation Learning and
  Beyond (GRL+ 2020)}, 2020.

\bibitem{ava_cascon}
S.A. Razi~Razavi et~al.
\newblock Generating labeled graphs using conditional {Wasserstein} {GAN}s.
\newblock In {\em 35th International Conference on Collaborative Advances in
  Software and Computing (CASCON)}. IEEE Computer Society, 2025.

\bibitem{ava_arxiv}
S.A. Razi~Razavi et~al.
\newblock Adaptive edge learning for density-aware graph generation.
\newblock {\em arXiv preprint arXiv:2601.23052}, 2026.

\bibitem{simonovsky2018graphvae}
M.~Simonovsky and N.~Komodakis.
\newblock {G}raph{VAE}: Towards generation of small graphs using variational
  autoencoders.
\newblock In {\em Intl. Conf. on Artificial Neural Networks}, 2018.

\bibitem{wang2018graphgan}
H.~Wang et~al.
\newblock {G}raph{GAN}: Graph representation learning with generative
  adversarial nets.
\newblock In {\em Proceedings of the AAAI Conference on Artificial
  Intelligence}, volume~32, 2018.

\bibitem{wang2024graph}
P.~Wang et~al.
\newblock Graph generative adversarial networks with evolutionary algorithm.
\newblock {\em Applied Soft Computing}, 164:111981, 2024.

\bibitem{you2018graph}
J.~You et~al.
\newblock Graph convolutional policy network for goal-directed molecular graph
  generation.
\newblock In {\em Advances in Neural Information Processing Systems}, 2018.

\end{thebibliography}

\end{document}